\documentclass{article}

\usepackage[preprint]{neurips_2025}

\usepackage[utf8]{inputenc} 
\usepackage[T1]{fontenc}    
\usepackage{hyperref}       
\usepackage{url}            
\usepackage{booktabs}       
\usepackage{amsfonts}       
\usepackage{subcaption}     
\usepackage{amsfonts}       
\usepackage{nicefrac}       
\usepackage{microtype}      
\usepackage{xcolor}         
\usepackage{amsmath}        
\usepackage{tabularray}
\usepackage{algorithm}
\usepackage{algorithmic}
\usepackage{graphicx}
\usepackage{amsfonts}       
\usepackage{newunicodechar}

\usepackage{wrapfig}
\usepackage{pifont}
\definecolor{mygray}{gray}{0.9}

\newunicodechar{，}{,}
\usepackage{natbib} 

\title{Marrying Autoregressive Transformer and Diffusion with Multi-Reference Autoregression}

\author{%
  Dingcheng Zhen\textsuperscript{\rm 1}$^\ast$, Qian Qiao\textsuperscript{\rm 1}$^\ast$, Xu Zheng\textsuperscript{\rm 3}$^\ast$, Tan Yu\textsuperscript{\rm 1}$^\ast$, Kangxi Wu\textsuperscript{\rm 1,2}\thanks{Equal contribution. $\dagger$ Corresponding author, Shunshun Yin is the Project Leader} \\
  {\bfseries Ziwei Zhang, Siyuan Liu, Shunshun Yin}\textsuperscript{\dag}, {\bfseries Ming Tao} \\
  {\bfseries \textsuperscript{\rm 1}Soul AI, \textsuperscript{\rm 2}ICT, Chinese Academy of Sciences, \textsuperscript{\rm 3}HKUST (GZ)}\\
  \texttt{\small\{dingchengzhen, qiaoqian, yutan, wukangxi, yinshunshun, ming\}} \\
  \texttt{\small@soulapp.cn} \\
  \url{https://github.com/TransDiff/TransDiff}   
}

\begin{document}

\maketitle

\begin{figure}[ht]
\vspace{-26pt}
\begin{center}
    \includegraphics[width=0.99\linewidth]{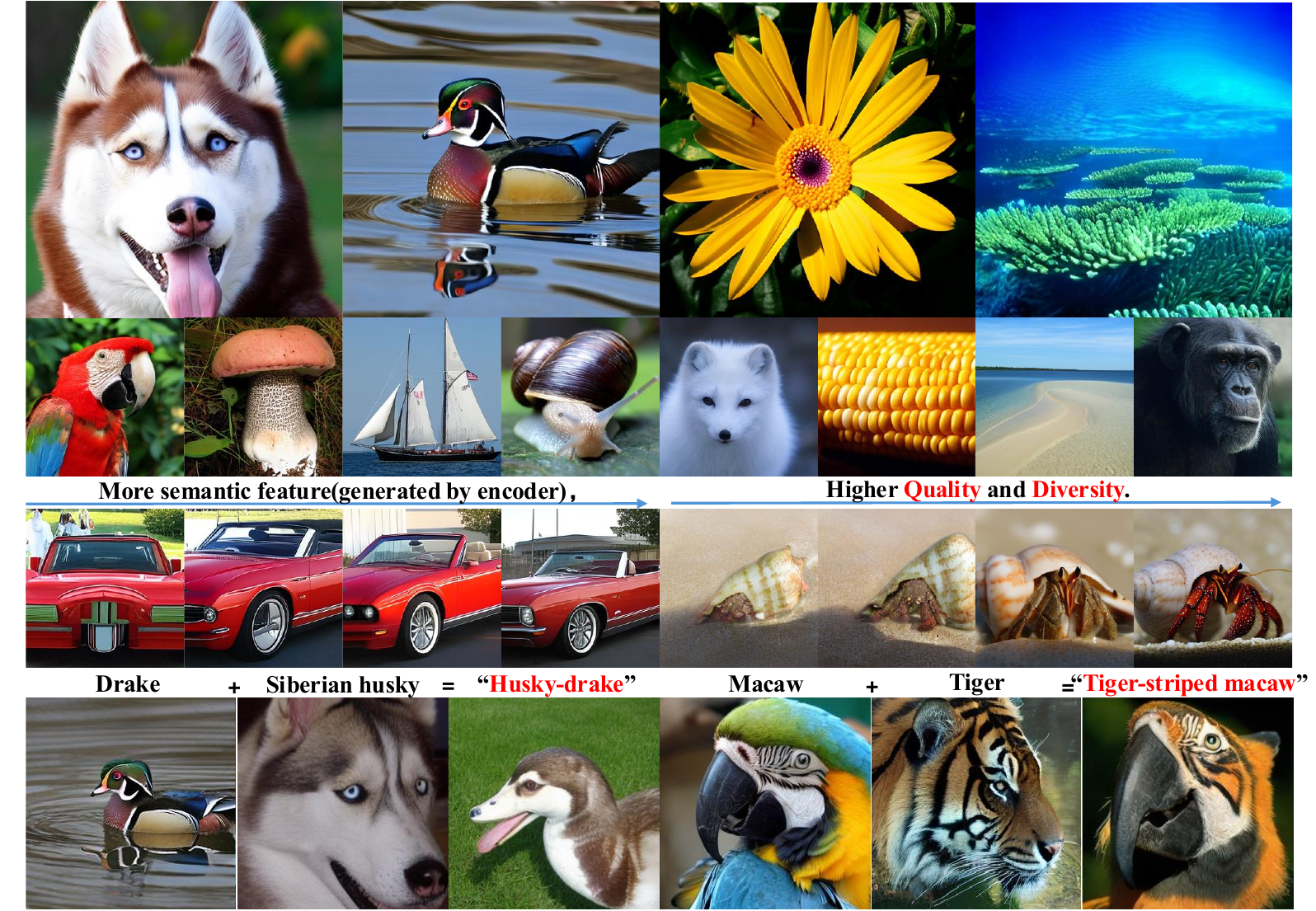}
\end{center}
\caption{\textbf{Generated samples from TransDiff trained on ImageNet.} \textbf{Top:} $512\times 512$ and $256\times 256$ samples. 
   \textbf{Middle:} effect of semantic feature diversity in TransDiff on image quality(left to right: increasing diversity.). 
   \textbf{Bottom:} results of semantic features fusion. (The first two columns show images from two classes; the third shows the fused result.)
   }
\label{fig1: abs}
\end{figure}

\begin{abstract}
We present TransDiff, the \textbf{\textit{first}} image generation model that marries Autoregressive (AR) Transformer with diffusion models. In this unified framework, TransDiff encodes labels and images into high-level semantic features and employs a diffusion model to estimate the distribution of image samples. On the ImageNet $256\times 256$ benchmark, TransDiff significantly outperforms all other image generation models based on standalone AR Transformer or diffusion models. Specifically, TransDiff achieves a Fréchet Inception Distance (FID) of \textbf{1.61} and an Inception Score (IS) of \textbf{293.4}, and further provides \textbf{$\times 2$} faster inference latency compared to state-of-the-art methods based on AR Transformer and \textbf{$\times 112$} faster inference compared to diffusion-only models. Moreover, building on the TransDiff model, we introduce a novel image generation paradigm, namely Multi-Reference Autoregression (MRAR), which performs autoregressive generation by predicting the next image. It enables the model to reference multiple previously generated images, thereby facilitating the learning of more diverse representations and improving the quality of generated images in subsequent iterations. By applying MRAR, the performance of TransDiff is improved, with the FID reduced from $\textbf{1.61}$ to $\textbf{1.42}$. We expect TransDiff to open up a new frontier in the field of image generation.
\begin{figure*}
     \centering
     \includegraphics[width=\linewidth]{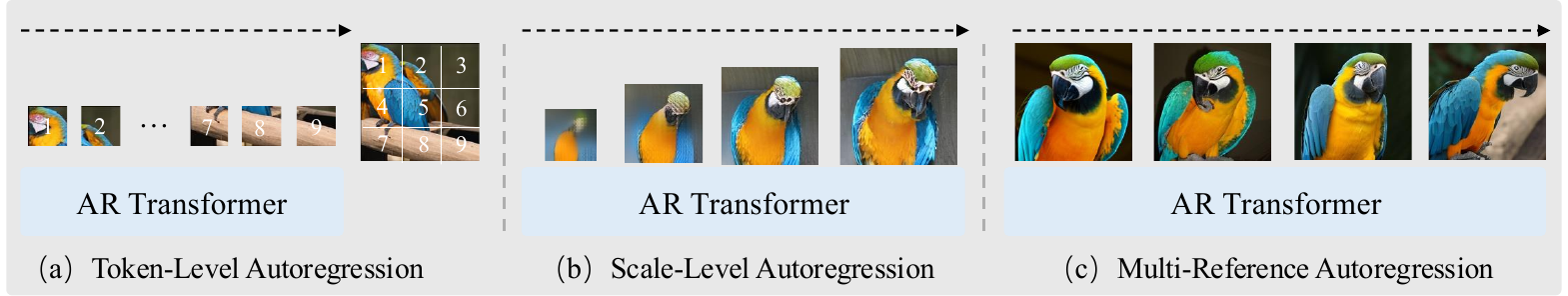}
    \caption{Comparison of Token-Level AR, Scale-Level AR and MRAR paradigms.}
    \label{fig: intro}
\end{figure*}
\end{abstract}
\section{Introduction}

Image generation has achieved remarkable progress driven by transformer architectures~\cite{esser2021taming,van2017neural,razavi2019generating} and diffusion models~\cite{ho2020denoising,dhariwal2021diffusion,nichol2021improved,peebles2023scalable,rombach2022high,chen2024pixart,flux2024}. Currently, two dominant approaches prevail: Autoregressive (AR) Transformers and diffusion models. AR Transformers employ vector quantization (VQ)~\cite{van2017neural} to convert images into discrete tokens for sequential generation. Diffusion models progressively denoise from random Gaussian noise to generate high-fidelity images.

However, both paradigms face fundamental limitations. AR Transformers suffer from the $O(n^2)$ computational complexity of causal attention. This complexity becomes prohibitive for high-resolution images. To mitigate this, most AR-based methods~\cite{tian2024visual,sun2024autoregressive} compress images into discrete latent representations (\emph{e.g.}, $32\times 32$ instead of $256\times 256\times 3$). This compression achieves computational efficiency at the cost of information loss through quantization. The discrete representation bottleneck fundamentally constrains generation quality. Conversely, diffusion models excel at producing high-quality images by modeling arbitrary probability distributions~\cite{li2024autoregressive}. However, their iterative denoising process results in significantly slower inference speeds compared to AR approaches.

A natural question arises: \textit{Can we take advantages of both models and have a new paradigm for image generation?} To this end, we present \textbf{TransDiff}, the first unified framework that combines AR Transformer with diffusion models. Different from the recent attempt BLIP3-o\cite{chen2025blip3}, which relies on two-stage training and sequentially learns AR transformer and diffusion model, our \textbf{TransDiff} is jointly trained with an end-to-end manner. Specifically, TransDiff employs an AR Transformer to encode inputs into high-level semantic features and apply a diffusion model as a decoder to generate images from these semantic representations. Through training, the AR transformer learns to extract precise semantic features. The diffusion decoder develops the capability to interpret these features effectively (as shown in Figure~\ref{fig1: abs} Middle and Bottom).

Building upon our \textbf{TransDiff}, we further investigate how to improve the AR paradigm and achieve better performance. Current autoregressive paradigms can be categorized into two types: token-level AR models~\cite{ramesh2021zero,cao2023efficient,chen2020generative,van2017neural,van2016conditional} and scale-level AR models~\cite{tian2024visual,liu2024alleviating,ma2024star,zhang2024var}. Token-level models generate images by predicting one token at each step. During generation, they observe only incomplete image tokens. Token-level models observe only incomplete image tokens during generation and scale-level models refer to coarse and blurry image versions at each step. Both paradigms rely on incomplete or imprecise information. This limitation constrains generation quality and diversity.

To bridge this gap, we furthre introduce \textbf{Multi-Reference Autoregression (MRAR)}. This is a novel AR generation paradigm where the model predicts a complete image at each step. Then MRAR enables the model to leverage comprehensive reference information from the previous steps during generation, achieving more diverse and higher-quality outputs (\emph{e.g.}, Figure~\ref{fig: intro}).

We validate TransDiff with MRAR on the challenging ImageNet~\cite{deng2009imagenet} benchmark. Our TransDiff achieves a \textbf{1.42} Fréchet Inception Distance (FID) score. This outperforms diffusion models with equal parameter count (FID 1.58). Meanwhile TransDiff also maintains inference speed comparable to state-of-the-art AR Transformer models. It significantly outperforms diffusion models in inference efficiency.
\section{Related Work}
\subsection{Diffusion Models for Image Generation}
Diffusion models gradually remove noise from images to restore them to clear images. Early works used U-Net as the denoiser~\cite{HoSCFNS22,RombachBLEO22,ho2020denoising,DBLP:conf/iclr/LiuG023,DBLP:conf/iclr/0011SKKEP21}. Such as, CDM~\cite{HoSCFNS22} uses cascaded U-net diffusion models to generate high-fidelity images. And, LDM~\cite{RombachBLEO22} performs the diffusion processes in the latent space of pretrained autoencoders and significantly reduces the computational cost. Recently, with the significant progress of the transformer architecture in the field of NLP, a growing number of works have begun using transformers as denoisers for image generation~\cite{PeeblesX23, DBLP:conf/icml/EsserKBEMSLLSBP24, DBLP:journals/corr/abs-2405-08748,flux2024}, such as, DiT~\cite{PeeblesX23} is the first work to use the transformer architecture instead of the previous U-Net architecture and to explore the scalability of diffusion transformers. And, Esser et al.~\cite{DBLP:conf/icml/EsserKBEMSLLSBP24} improved rectified flow models with perceptually-biased noise sampling and introduces a novel dual-modality transformer.

Despite the significant progress of diffusion models in the field of image generation, they often require higher resource consumption and longer inference times to achieve better results. In contrast, our approach consumes fewer resources and reduces inference time, while also achieving superior performance compared to diffusion models with the same number of parameters.
 
\subsection{Autogressive Models for Image Generation}
AR models generate images by regressively predicting the next token~\cite{LeeKKCH22, ChangZJLF22,DBLP:journals/corr/abs-2502-20388,DBLP:journals/corr/abs-2411-00776,esser2021taming} or next scale~\cite{tian2024visual,liu2024alleviating,ma2024star,zhang2024var}. 
For next token prediction, RQ-Transformer~\cite{LeeKKCH22} utilizes a Residual-Quantized VAE (RQ-VAE) for autoregressive image generation, resulting in lower computational costs and faster sampling speeds. Maskgit~\cite{ChangZJLF22} utilizes a bidirectional transformer decoder that learns to predict randomly masked image tokens by attending to tokens in all directions, significantly enhancing the efficiency of autoregressive decoding. For next scale prediction,  VAR~\cite{tian2024visual} proposes a coarse-to-fine "next-scale prediction" approach to redefine autoregressive learning for images, enabling GPT-style AR models to surpass diffusion transformers in image generation for the first time. DiMR~\cite{liu2024alleviating}
incrementally enhances features at various scales, enabling a gradual improvement in visual detail from coarse to fine levels.

More recently, BLIP3-o\cite{chen2025blip3} relies on two-stage training process: it first uses EVA-CLIP to encode images into continuous visual embeddings, which are then reconstructed using a diffusion model. Then the AR Transformer in BLIP3-o is trained to generate these visual embeddings. 
However, the EVA-CLIP's original training objective is not aligned with generation tasks, thus the visual embeddings are not optimal for the diffusion models.

Different from all the above studies, this paper proposes Multi-Reference Autoregression (MRAR), which is a novel autoregressive generation paradigm. It predicts a complete image each step, allowing the model to refer to more comprehensive and complete information during prediction, thereby generating higher-quality and more diverse images.

\begin{figure*}[t]
     \centering
     \includegraphics[width=0.90\linewidth]{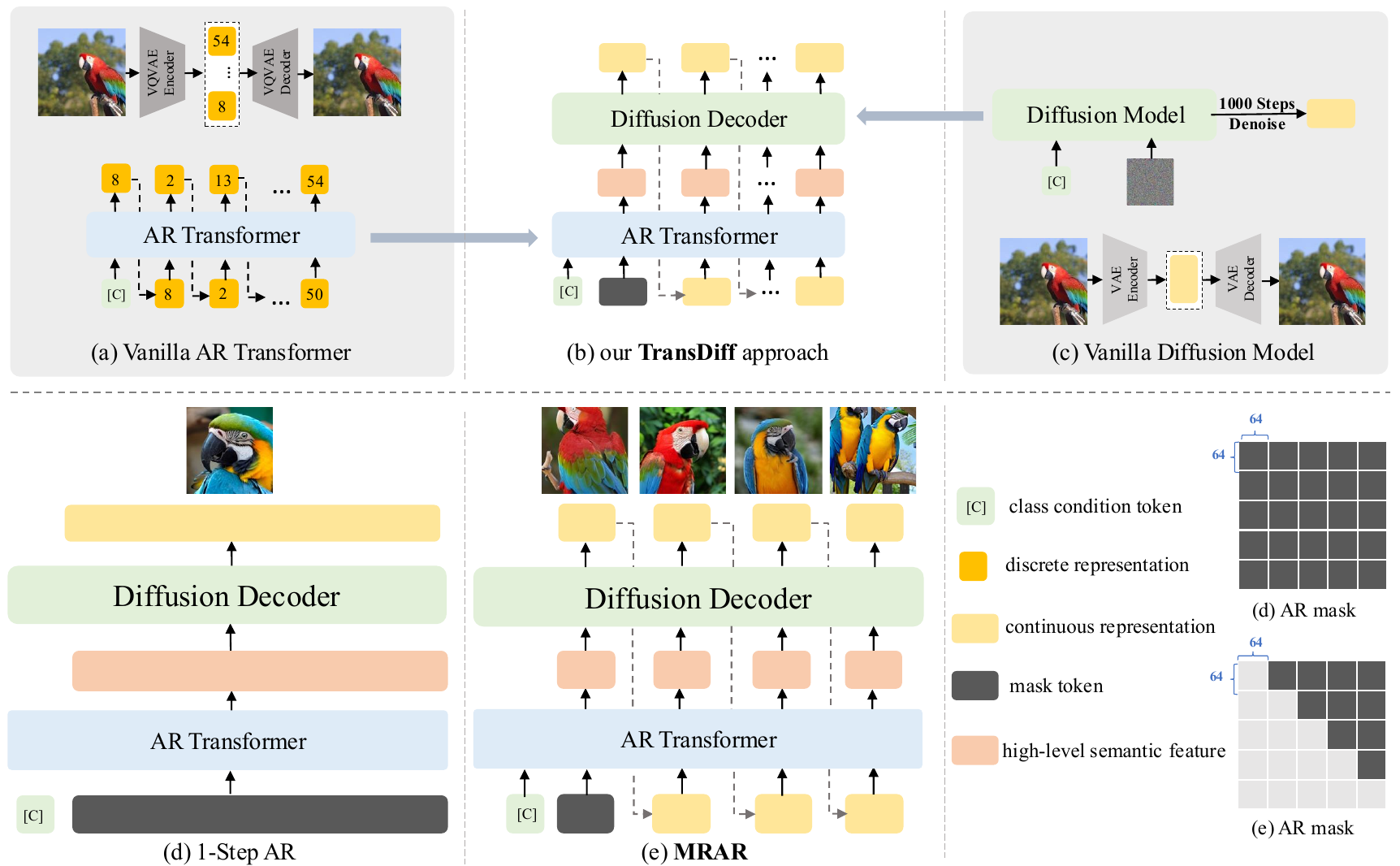}
    \caption{Overview of our Transdiff. Wherein (a) and (c) are the Vanilla Autoregressive Transformer and the Vanilla Diffusion Model, respectively. (b) is our TransDiff approach. (d) and (e) provide a detailed explanation of 1-step AR and MRAR.}
    \label{fig: TransDiff} 
\end{figure*}
\section{Methodology}

\subsection{TransDiff}
We introduce TransDiff, which first unifies two dominant methods in contemporary image generation: the AR Transformer and the diffusion model. In brief, we employ the AR Transformer to encode input into high-level semantic features serving as a conditioning signal, and utilize a diffusion decoder~\cite{peebles2023scalable} to generate the image. This section begins with a brief introduction of these two paradigms, followed by a detailed explanation of how we systematically integrate them into a unified framework.

\subsection{AR Transformer (ART)} 
Vanilla autoregressive image generation methods typically rely on a pre-trained tokenizer to quantize images into discrete tokens, which are then used to reconstruct the original image. However, the limited capacity of the VQ model imposes a fundamental constraint on the reconstruction quality
. This bottleneck arises from the inherent trade-off between compression ratio and information retention: a smaller compression ratio increases computational complexity, while a larger compression ratio leads to significant loss of information during quantization. 

Unlike common AR Transformer models~\cite{radford2019language,brown2020language}, which take discrete tokens as input, we let $x = (x_1, x_2, \ldots, x_N)$ denote a sequence of continuous features, where  $x_n\in \mathbb{R}^{l \times d}$ . The AR Transformer models assume that the probability of each $x_n$ depends exclusively on its preceding tokens $(x_1, x_2, \ldots, x_{n-1})$. This assumption of unidirectional dependency enables the joint probability of the entire sequence to be factorized as a product of conditional probabilities:
\begin{equation}
\setlength\abovedisplayskip{5pt}
\setlength\belowdisplayskip{5pt}
 p(x) = \prod_{n=1}^{N} p(x_n \mid x_1, x_2, \ldots, x_{n-1})  
\end{equation}
The loss function is typically defined as:

\begin{equation}
\mathcal{L}_{\textit{AR}}=\frac{1}{N}
\sum_{n=0}^{N}\mathrm{LF}(x_n,\text{ART}(x_0,x_1, ... ,x_{n-1}))
\label{loss:AR}
\end{equation}
where $\mathrm{LF}$ denotes the loss function(e.g., cross-entropy loss).

\subsection{Diffusion Decoder}
We adopt DiT~\cite{peebles2023scalable} as our diffusion decoder. In addition, we replace the original DDPM~\cite{ho2020denoising} objective with Flow Matching~\cite{DBLP:conf/iclr/LipmanCBNL23}. Flow Matching learns a transformation from latent variables $\epsilon \sim \mathcal{N}(0, 1)$ to samples drawn from the data distribution, formulated through an ordinary differential equation (ODE):
\begin{equation}
    \frac{dx^t}{dt} = \psi_{\theta}(x^t,\ t,\ c)
\end{equation}
Here, $\psi_{\theta}$ represents a learnable velocity field parameterized by the model weights $\theta$. $t \in [0, 1]$ denotes the continuous time variable and $x^t$ refers to the data point at time $t$. $c$ is a conditioning input.

To improve optimization efficiency, we adopt rectified flow~\cite{DBLP:conf/iclr/LiuG023}, which assumes that the trajectory connecting a data sample $x$ and a noise sample $\epsilon$ in latent space follows a straight-line path:
\begin{equation}
    x^t = (1 - t) \cdot x + t \cdot \epsilon
\end{equation}
where $\epsilon \sim \mathcal{N}(0, I)$ and $t \in [0, 1]$.
Then, the loss of the diffusion decoder can be expressed as:
\begin{equation}
\begin{split}
\mathcal{L}_{\text{diff}} =\ 
&\mathbb{E}_{\substack{t \sim \text{Uniform}([0, 1]) \\ x \sim P \\ c \sim \text{ART}}}
\left[
\left\|
(\epsilon - x) - \psi_{\theta}\left(x^t,\ t,\ c\right)
\right\|^2
\right]
\end{split}
\label{loss:diff}
\end{equation}
where $P$ represents the distribution of the training dataset, and $c$ refers to a condition of diffusion decoder generated by AR Transformer Encoder, based on both the image class and the known image.

\subsection{The Final TransDiff Architecture}
We present TransDiff, which serves as the first framework that enable joint training of the AR Transformer and the diffusion model(\emph{i.e.}, diffusion decoder). While AR Transformer-based image generation methods use VQ-VAE to map images into discrete tokens, our approach employs a VAE to project images into a continuous latent space. Subsequently, our AR Transformer encodes the obtained image latent into high-level semantic features. 
\begin{equation}
\begin{aligned}
        & c =\text{ART}(\text{input})
\end{aligned}
\end{equation}
where $\text{input} \in \mathbb{R}^{(h \times w) \times d}$ denotes the the continuous latent of image, $c\in \mathbb{R}^{(h / f\times w/f)\times (d \times f \times f)}$ denotes the condition for diffusion decoder, where $f$ denotes the compression ratio.
Then, these features serve as conditioning inputs to the diffusion decoder to generate the image. Detailed as follows:
\begin{equation}
\begin{aligned}
    &\text{output }= \text{Diff}(c,\epsilon)
\end{aligned}
\end{equation}
where  $\text{Diff}$ denotes the Diffusion Deocder,   $\epsilon \sim \mathcal{N}(0, I)$ and $output \in \mathbb{R}^{(h \times w) \times d}$ denotes the the continuous latent of output image.
Finally, by combining the loss of AR Transformer model (~\emph{i.e.}, Eq.~\ref{loss:AR} )and the diffusion decoder(~\emph{i.e.}, Eq.~\ref{loss:diff}), the overall loss function can be defined as:
\begin{equation}
\begin{aligned}
\mathcal{L}_{\text{all}} = \frac{1}{N}&
\sum_{n=0}^{N} 
\mathbb{E}_{\substack{t \sim \text{Uniform}([0, 1]), x \sim P}}
\Bigg[
\Big\| 
(\epsilon - x_n)-  \\
& \psi_{\theta}(x_n^t,\ t,\ \text{ART}(x_0, x_1, \dots, x_{n-1}))
\Big\|^2 
\Bigg]
\end{aligned}
\end{equation}
where $x_n$ is the $n$-th input latent for AR transformer, and $x_n^t$ represents the diffusion sample of $x_n$ at time step $t$.

\subsection{Multi-Reference Autoregressive (MRAR)}

In the AR component of TransDiff, we introduce a novel AR generation paradigm, Multi-Reference Autoregressive (\textbf{MRAR}), which diverges from traditional AR generation paradigm. Unlike the standard 1-step AR method, which generates predictions sequentially from the previous output, MRAR incorporates multiple references from the entire sequence of previous outputs. Our experimental results in Table~\ref{tab:MRAR} demonstrate that MRAR consistently outperforms the one-step AR strategy, offering significant improvements in performance. Based on these observations, we adopt MRAR as the preferred AR strategy in our model, positioning it as a new paradigm for AR modeling.

\subsubsection{1-Step AR}
As in the Figure~\ref{fig: TransDiff}(d), the goal of 1-Step AR encode the input high-level into semantic features $c$ through a single forward pass of the AR transformer, conditioned on the class label, which is then decoded by the diffusion decoder to generate the output. 
Specifically, the input is:
\begin{equation}
    \begin{aligned}
        &\text{Mask} = \text{Concat}( [\text{m},\text{m},...,\text{m}])\\
        &\text{input} = \text{Concat}([\text{C},\text{Mask}])
    \end{aligned}
    \label{process:AR}
\end{equation}
where, $\text{C}$ denotes the class embedding of image, and $\text{m}$ represents the mask embedding. And $\text{C} \in \mathbb{R}^{64 \times d}$, $\text{Mask}\in \mathbb{R}^{(h \times w) \times d}$. 
Then, we follow the forward propagation and loss computation of TransDiff:
\begin{equation}
    \begin{aligned}
        &c = \text{ART(input)}\\
        &\text{output} = \text{Diff}(c,\epsilon)
    \end{aligned}
    \label{process:AR}
\end{equation}
Furthermore, as shown in the Figure~\ref{fig: TransDiff} (d), in 1-Step AR, the attention mask of the AR Transformer in 1-Step AR is a bidirectional attention matrix filled with zeros.
The inference process of 1-Step AR is shown in the Appendix.

\subsubsection{The MRAR Paradigm}
In the 1-Step AR setting, we observe that in the image features $c$ (as defined in the Eq.~\ref{process:AR}) of a category, the more diverse the features relevant to that category, the better the quality of the generated images (A detailed experimental analysis is provided in the Figure.~\ref{fig:semantic_feature}). 
Based on the above observations, we propose MRAR, which learns more comprehensive features by referencing multiple generated images in the previous steps. This allows it to generate higher-quality images through iterations
. The detailed process is shown in the Figure~\ref{fig: TransDiff} (e).
Therefore, building on the training of 1-Step AR, we further fine-tune the model using MRAR. 
Specifically, the input can be formalized as follows:
\begin{equation}
    \begin{aligned}
        &\text{input} = \text{Concat}([\text{C},\text{Mask},x_{\text{img}_0},x_{\text{img}_1},...,x_{\text{img}_{n-1}}])
    \end{aligned}
\end{equation} 
where $\{x_{\text{img}_i}\}_{i=1}^{n} \in \mathbb{R}^{(h / f \times w / f) \times (d \times f \times f)}$ denotes the latent of the $i$-th image under the same label, and $f$ is the compression ratio. $\text{Mask} \in \mathbb{R}^{(h / f \times w / f) \times (d \times f \times f)}$ represents the mask embedding.
Then, we follow the forward propagation and loss computation of TransDiff:
\begin{equation}
    \begin{aligned}
        &c = \text{ART(input)}=\text{Concat}([c_{\text{img}_0}, c_{\text{img}_1},c_{\text{img}_2},...,c_{\text{img}_n}])\\
        &\text{output} = \text{Diff}(c,\epsilon) = \text{Concat}([o_{\text{img}_0}, o_{\text{img}_1},o_{\text{img}_2},...,o_{\text{img}_n}])
    \end{aligned}
\end{equation} 
where $\{c_{\text{img}_i}\}_{i=0}^{n} \in \mathbb{R}^{(h / f \times w / f) \times (d \times f \times f)}$ and $\{o_{\text{img}_i}\}_{i=0}^{n} \in \mathbb{R}^{(h \times w) \times d}$ denote the condition features and the final output of the $i$-th image, respectively. Moreover, the final generated image is denoted as $o_{\text{img}_n} \in \mathbb{R}^{(h \times w) \times d}$. As illustrated in the Figure~\ref{fig: TransDiff} (e), the attention mask in the AR Transformer of MRAR is implemented as a causal attention matrix.
The inference process of MRAR is shown in the Appendix.
\section{Experiments}
We demonstrate through a series of experiments that TransDiff is a efficient method capable of fitting image distributions. We validate that the novel MRAR image generation paradigm effectively enriches TransDiff's feature representations, which enhances the diversity of generated images. Specifically, we investigate the following questions:

\textbf{(I)} Can TransDiff beat diffusion-only and transformer-only approaches? (Table~\ref{tab:TransDiff beats others2})

\textbf{(II)} Does the AR Transformer in TransDiff extract high-level semantic features, and can the diffusion decoder interpret these features to generate high-quality images? (Figure~\ref{fig:semantic_feature}, Figure~\ref{fig1: abs})

\textbf{(III)} Is the MRAR paradigm superior to Token-Level AR and Scale-Level AR paradigms? ( Table~\ref{tab:MRAR}, Figure~\ref{fig:humaneval})

\subsection{Experimental Setup}
\noindent \textbf{Datasets and Metrics} We conduct comprehensive experiments on the ImageNet~\cite{deng2009imagenet} at $256\times 256$ and $512 \times 512$ resolution, evaluating our method through quantitative metrics including Fréchet Inception Distance (FID)~\cite{heusel2017gans}, Inception Score (IS)~\cite{salimans2016improved}, Precision-Recall analysis and computational efficiency (inference time). 
The Euler–Maruyama method, presented in the Appendix, is employed during the inference stage.
The implementation details are shown in Appendix.

\subsection{Results and Analysis}
\subsubsection{Comparison with diffusion-only and AR Transformer-only methods}
In the Table~\ref{tab:TransDiff beats others2}, we compare the performance of three architectures among ours, diffusion only, and the AR only methods. Under similar parameter settings, TransDiff-L with MRAR achieve the \textbf{lowest} FID of \textbf{1.49} and TransDiff with 1-step AR achieve a competitive remarkable speed of \textbf{0.2s/image}. 

\textbf{\ding{172} Diffusion Decoder \textit{vs.} Diffusion Loss.} 
We assess the effectiveness of our usage of diffusion decoder against the the diffusion loss in MAR~\cite{li2024autoregressive}. 
Specifically, we keep the AR Transformer unchanged and perform a comparative evaluation of the differences in FID, IS, and inference time (Time) for a single image across varying inference steps between the diffusion loss (MAR) and the diffusion decoder (TransDiff). In the Table~\ref{tab:DiffdecvsDiffloss}, our method, which incorporates a diffusion decoder, demonstrates superior performance by reducing the FID from \textbf{1.78 to 1.49} and achieving faster inference. At the same time, it is important to note that our approach has the \textbf{capability for one-step inference}, whereas MAR which relies on diffusion loss does not.

\textbf{\ding{173} Comparison between TarnsDiff and MAR.} As in Figure\ref{fig:feature_mix}, it is obvious that MAR's encoder fails to extract the semantic features while our TransDiff's encoder exactly learn the semantic within the input images. For example, we can fuse semantic features from Icecream and Pug(128 features from each), generate ‘Pug looked icecream’. 

\begin{table*}[t]
\centering
\scalebox{0.85}{
\begin{tblr}{
  cells = {c},
  cell{1}{1} = {r=2}{},
  cell{1}{2} = {r=2}{},
  cell{1}{3} = {r=2}{},
  cell{1}{4} = {c=2}{},
  cell{1}{6} = {c=4}{},
  cell{1}{10} = {r=2}{},
  vline{2-4,6,10} = {1-32}{0.05em},
  hline{1,29} = {-}{0.20em},
  hline{3,6,9,12,20,23} = {-}{0.05em},
}
\textbf{Type} & \textbf{Model} & \textbf{\#Params} & \textbf{w/o CFG} & & \textbf{w/ CFG} & & & & \textbf{Time} $\downarrow$\\ 
              &                &                   & \textbf{FID} $\downarrow$ & \textbf{IS} $\uparrow$ & \textbf{FID} $\downarrow$ & \textbf{IS} $\uparrow$ & \textbf{Pre.} $\uparrow$ & \textbf{Rec.} $\uparrow$ & \\
GAN           & BigGAN~\cite{brock2018large}                    & 112M              & 6.95             & 224.5        & -               & -           & -    & -             & -             \\
GAN           & GigaGAN~\cite{kang2023scaling}                   & 569M              & 3.45             & 225.5       & -               & -           & -    & -             & -             \\
GAN           & StyleGan-XL~\cite{sauer2022stylegan}               & 166M              & 2.30             & \textbf{265.1}     & -               & -           & -    & -             & 0.3           \\
Mask.         & MaskGIT~\cite{chang2022maskgit}                   & 227M              & 6.18             & 182.1      & -               & -           & -    & -             & 0.5           \\
Mask.         & MAGE~\cite{li2023mage}                      & 230M              & 6.93             & 195.8        & -               & -           & -    & -             & -             \\
Mask.         & MAGVIT-v2~\cite{yu2023language}                 & 307M              & 3.65             & 200.5      & 1.78            & \textbf{319.4}     & - & -          & -             \\
Diffusion     & LDM-4~\cite{rombach2022high}                     & 400M              & 10.56            & 103.5       & 3.60            & 247.7       & \textbf{0.87} & 0.48          & -             \\
Diffusion     & DiT-XL/2~\cite{peebles2023scalable}                  & 675M              & 9.62             & 121.5      & 2.27            & 278.2       & 0.83 & 0.57          & 45            \\
Diffusion     & DiffT~\cite{hatamizadeh2024diffit}                     & -                 & -                & -           & 1.73            & 276.5       & 0.80 & \textbf{0.62}          & -             \\
AR            & VQGAN~\cite{esser2021taming}                     & 227M              & 18.65            & 80.4        & -               & -           & -    & -             & 19            \\
AR            & VQGAN~\cite{esser2021taming}                     & 1.4B              & 15.78            & 74.3         & -               & -           & -    & -             & 24            \\
AR            & LlamaGen-3B~\cite{sun2024autoregressive}               & 3.1B              & 9.95             & 112.87       & 2.18               & 263.3           & 0.81    & 0.58             & -             \\
AR            & VAR-d16~\cite{tian2024visual}                   & 310M              & -                & -          & 3.30            & 274.4       & 0.84 & 0.51          & 0.4           \\
AR            & VAR-d20~\cite{tian2024visual}                   & 600M              & -             & -             & 2.57            & 302.6       & 0.83 & 0.56          & 0.5           \\
AR            & VAR-d24~\cite{tian2024visual}                   & 1.0B              & -                & -           & 2.09            & 312.9       & 0.82 & 0.59          & 1             \\
AR            & RAR-L~\cite{DBLP:journals/corr/abs-2411-00776}                     & 461M              & -                 & -             & 1.70            & 290.5       & 0.82 & 0.60          & 15.0             \\
AR            & RAR-XL~\cite{DBLP:journals/corr/abs-2411-00776}                    & 955M              & -                & -             & 1.50            & 306.9       & 0.80 & \textbf{0.62}     & 8.3             \\
MAR           & MAR-B~\cite{li2024autoregressive}                     & 208M              & 3.48             & 192.4       & 2.31          & 281.7       & 0.82 & 0.57          &  0.5             \\
MAR           & MAR-L~\cite{li2024autoregressive}                     & 479M              & 2.60             & 221.4       & 1.78            & 296.0       & 0.81 & 0.60          & 1.1            \\
MAR           & MAR-H~\cite{li2024autoregressive}                     & 943M              & 2.35             & 227.8       & 1.55            & 303.7       & 0.81 & \textbf{0.62}          & 2.4             \\
TransDiff  & TransDiff-B, 1-Step AR &      290M         &   5.09          &  153.5         &     2.47         &   244.2     &    0.81  &      0.56         &       0.1        \\
TransDiff  & TransDiff-B, MRAR      &      290M         &   5.56        &   148.1  &       2.25          &    244.3         &   0.80   &     0.57          &     0.4          \\
TransDiff  & TransDiff-L, 1-Step AR &      683M         &    3.05       &   185.7 &    1.69        &    282.0      &    0.81  &        0.60       &     0.2        \\
TransDiff  & TransDiff-L, MRAR      &     683M          &    2.95          &  192.7           &     1.49            &   282.2          &  0.82    &       0.60        &    0.8        \\
TransDiff  & TransDiff-H, 1-Step AR &      1.3B         & \textbf{2.23} &   210.1  &      1.61           &    293.4         &   0.81   &     0.61          &           0.4    \\
TransDiff  & TransDiff-H, MRAR      &     1.3B         &  2.50          &  220.3    &     \textbf{1.42}          &      301.2       &   0.81   &     \textbf{0.62}          &               1.6  
\end{tblr}
}
\caption{\textbf{Comprehensive comparative analysis of various model architectures on the class-conditional ImageNet 256×256.} Metrics include FID$\downarrow$, IS$\uparrow$, precision (Pre.$\uparrow$), recall (Rec.$\uparrow$), and per\text{-}image generation time (Time$\downarrow$). Columns "w/o CFG" and "w/ CFG" denote results with and without classifier-free guidance. }
\label{tab:main}
\vspace{-14pt}
\end{table*}
\begin{table*}[t]
\centering
    \begin{tabular}{@{}lrlcc@{}}
        \toprule
        \textbf{Type} & \textbf{Model} & \textbf{Params} & \textbf{FID} & \textbf{IS} \\ 
        \midrule
        Diffusion & DiT-XL/2 & 675M & 2.27 & 278.2 \\
        Diffusion &  DiffT & 676M & 1.73 & 276.5 \\ 
        \midrule
        AR & RAR-L & 461M & 1.70 & 290.5 \\
        AR & VAR-d20 & 600M & 2.57 & \textbf{302.6} \\ 
        \midrule
        Hybrid & TransDiff-L 1-Step AR & 683M & 1.69 & 282.0\\
        Hybrid & TransDiff-L MRAR & 683M & \textbf{1.49}& 282.2\\ 
        \bottomrule
    \end{tabular}
    \caption{Comparison on ImageNet $256\times 256$ with existing diffusion-only and transformer-only approaches.}
    \label{tab:TransDiff beats others2}
\end{table*}

\begin{table*}[t]
\centering
    \begin{tabular}{@{}lcccc@{}}
        \toprule
        \textbf{Model} & \textbf{Steps} & \textbf{FID} $\downarrow$ & \textbf{IS} $\uparrow$ & \textbf{Time} \\ 
        \midrule
        MAR-L & 1 & 336.58 & 1.3 & - \\
        MAR-L & 256 & 1.78 & \textbf{296.0} & 1.1s \\ 
        \midrule
        TransDiff-L 1-Step AR & 1 & 1.69 & 282.0& 0.2s \\
        TransDiff-L MRAR & 4 & \textbf{1.49}& 282.2 & 0.8s \\ 
        \bottomrule
    \end{tabular}
    \caption{Quantitative comparison on ImageNet $256\times 256$ between Diffusion Decoder and Diffusion Loss. It is important to note that our approach has the \textbf{capability for one-step inference}, \textbf{whereas MAR does not}.}
    \label{tab:DiffdecvsDiffloss}
\end{table*}

\textbf{\ding{174} Comparison with SoTA Methods.} 
We evaluate TransDiff models of different sizes (Base, Large, and Huge) on the ImageNet $256\times 256$ conditional generation benchmark, and systematically compare them with state-of-the-art image generation model families. As quantitatively demonstrated in the Table~\ref{tab:main}, TransDiff achieve exceptional performance metrics characterized by extraordinarily \textbf{lowest FID (e.g., 1.42)} and competitive \textbf{IS (e.g., 301.2)}. These results comprehensively validate the architectural innovation of TransDiff in simultaneously advancing both generation quality and computational efficiency, surpassing existing diffusion-based models as well as autoregressive/masked autoregressive frameworks. Additionally, these advantages hold true on the $512\times 512$ synthesis benchmark (\textbf{lowest FID of 2.51}), which is provided in the Appendix.

\subsubsection{AR Transformer’s Representation}
\label{sec:representation}
To validate our hypothesis that more diverse image semantic features lead to higher-quality generated images, 
we first introduce a novel,independent metric for evaluating representational diversity by quantifying the L2 norm of cosine similarity matrix constructed from the 256 semantic representations for a single image. More information is shown in Appendix.

\textbf{\ding{172} Quantitative Analysis of Image Semantic Features.} As in the Figure~\ref{fig:semantic_feature}, 
the results demonstrate that as the number of training steps increases, the FID score consistently decreases while image diversity simultaneously improves, indicating enhanced image quality. This correlation suggests that prolonged training not only refines feature representations but also promotes greater diversity in generated samples, leading to higher-quality image generation.

\begin{figure}[t]
\centering
    \includegraphics[width=1.0\linewidth]{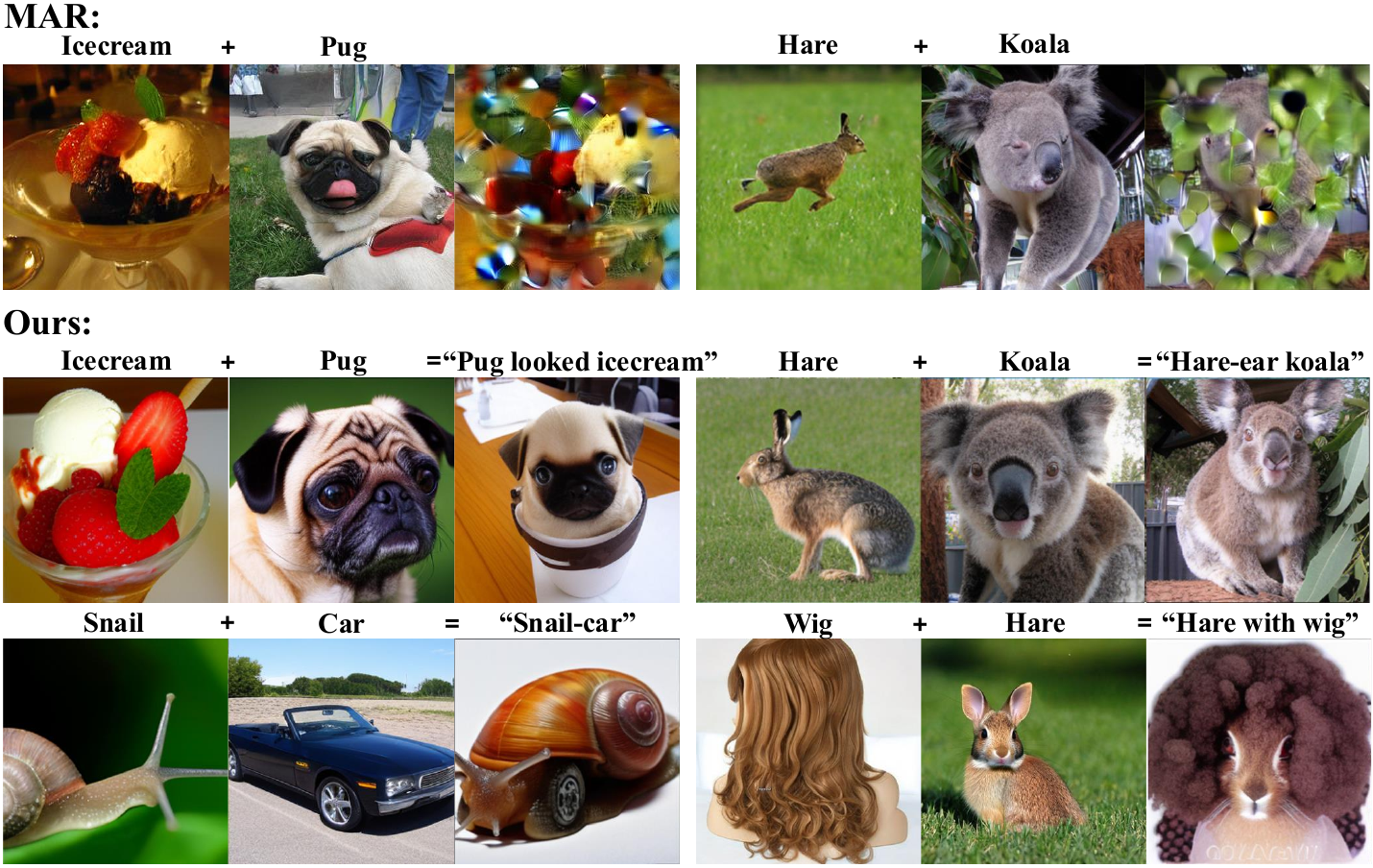}
\caption{
Results of semantic features fusion from images of different classes. (The first two columns show images from two classes; the third shows the fused result.)}
\label{fig:feature_mix}
\end{figure}

\begin{figure}[t]
    \centering
            \includegraphics[width=1.0\linewidth]{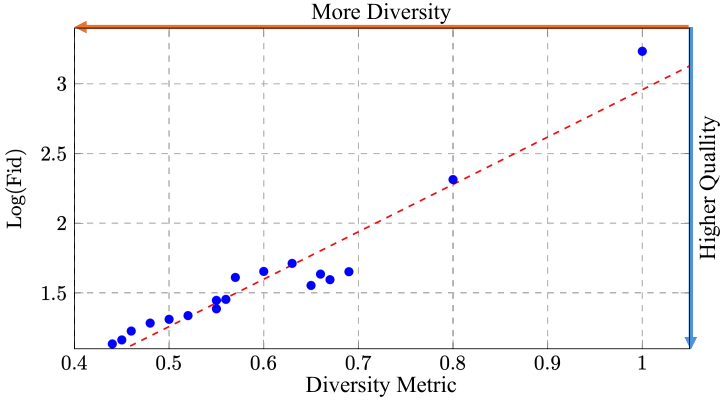}
            \caption{The \textbf{FID score} consistently decreases while \textbf{Diversity Metric} among generated samples drops, indicating improved image quality and diversity. }
            \label{fig:semantic_feature}
\end{figure}

\textbf{\ding{173} Qualitative Analysis of Image Semantic Features.}
The qualitative analysis is presented in the Figure~\ref{fig1: abs} . In the Middle, the results generated by the diffusion decoder using $32$, $64$, $128$, and $256$ image semantic features are shown. It can be observed that as the number of image features increases, the generated images exhibit richer details and improved visual quality, which further validates our conclusions. In addition, bottom of the Figure~\ref{fig1: abs} and  Figure~\ref{fig:feature_mix} reveals an interesting phenomenon. When we fuse semantic features (128 features from each) from images of different classes (\emph{e.g.}, Macaw and Tiger), generated images exhibit characteristics of both original classes (\emph{e.g.}, "Tiger-striped macaw"). This observation demonstrates that features obtained through our AR Transformer indeed capture high-level semantic features of images.

\subsubsection{MRAR beats Token AR and Scale-Level AR paradigms}
Based on the experimental analysis in AR Transformer’s representation, we observe that a higher diversity measure corresponds to a lower FID score, indicating better image quality. Thus, we propose the MRAR paradigm to enable the incorporation of more comprehensive representations.

\begin{table}[t]
    \centering
    \begin{tabular}{c|c|c|c}
        \hline
        \textbf{Model} & \textbf{Measure} $\downarrow$ & \textbf{FID} $\downarrow$ & \textbf{IS} $\uparrow$ \\ 
        \hline
        TransDiff-L, 1-step AR & 0.44 & 1.69 & 282.0 \\
        TransDiff-L, Scale AR  & 0.64 & 2.78 & 279.8 \\
        TransDiff-L, MRAR      & 0.39 & \textbf{1.49}& \textbf{282.2} \\
        \hline
    \end{tabular}%
    \caption{\textbf{MRAR \textit{vs.} scale AR.} Compared to the baseline, MRAR achieves a \textbf{higher diversity} (lower measure), resulting in a decrease in FID from \textbf{1.69 to 1.49}. In contrast, Scale-Level AR significantly reduces the diversity.}
    \label{tab:MRAR}
\end{table}
\begin{figure}[t]
    \centering
            \includegraphics[width=\linewidth]{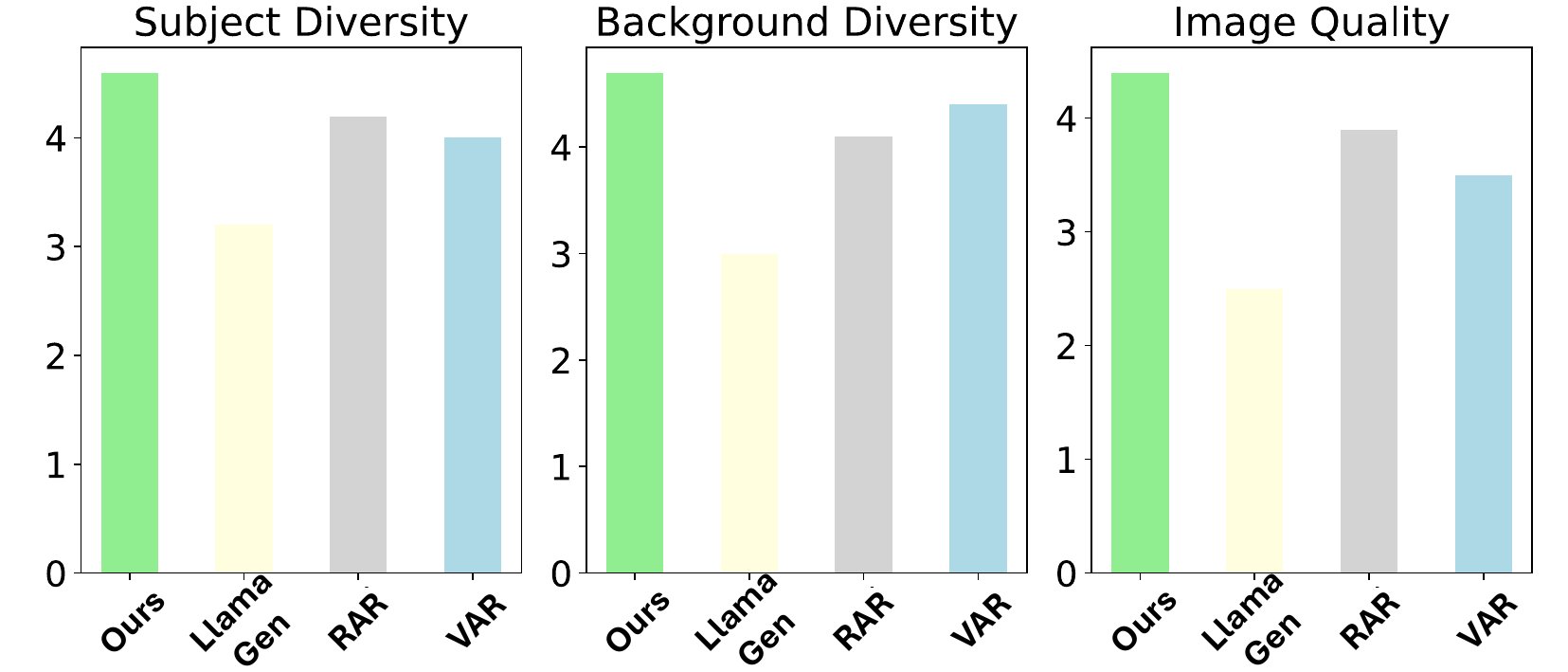}
            \caption{Human evaluation on our TransDiff and other SoTA methods of Token-Level AR and Scale-Level AR.}
            \label{fig:humaneval}
\end{figure}
\textbf{\ding{172} Human Evaluation on MRAR, Token-Level AR and Scale-Level AR.} 
To verify the superiority of our MRAR paradigm compared to token-level (\emph{e.g.}, LlamaGen-3B and RAR-XL) and scale-level AR paradigms (\emph{e.g.}, VAR-d24), we conducted a human evaluation. Sixty participants ($66.7\%$ aged $24-30$, $33.3\%$ aged $30-40$; $50\%$ male, $50\%$ female; $83.3\%$ with AIGC model experience) evaluate each image using a 5-point Likert scale across three dimensions: subject diversity, background diversity, and image quality. A total of $200$ images are presented in random order to minimize bias and better capture subjective perceptions of visual diversity and quality. As shown in the Figure~\ref{fig:humaneval}, participants consistently rated the MRAR paradigm higher than the token-level and scale-level AR paradigms in all three dimensions.

\textbf{\ding{173} Ablation on MRAR \textit{vs.} Scale-Level AR.}
We fine-tune the model using both the scale-level AR\cite{tian2024visual} and MRAR paradigms based on 1-Step AR training. The results in Table~\ref{tab:MRAR} show that, compared to 1-Step AR, MRAR achieves a higher measure, indicating greater image feature diversity, and a lower FID score, signifying better image generation quality. In contrast, the scale-level AR paradigm shows a lower measure, indicating reduced image feature diversity, and a higher FID score, reflecting poorer generation quality. This demonstrates that MRAR, by referencing more complete images, captures more diverse semantic features, leading to improved results.
\section{Conclusion and Limitations}
\textbf{Conclusion:} In this paper, we proposed TransDiff, demonstrating through strong experiments the significant potential of marrying AR Transformers with diffusion models, without being constrained by discrete representations. Notably, our proposed MRAR paradigm showcases excellent performance, opening up a new direction in autoregressive image generation. We hope this new paradigm can inspire future research in both image and video generation. 

\textbf{Training Data Constraints:} Primarily trained on ImageNet, the model lacks the scale and diversity of commercial proprietary datasets. This limits its ability to generate high-resolution, photorealistic images with complex semantic details, where visual quality is notably inferior to industrial models. Future work should explore larger, more diverse datasets to address this discrepancy.

\bibliographystyle{plain}
\newpage
\bibliography{References}

\appendix

\newpage 
\appendix

\section{Appendix}

\section{Inference Process of 1-Step AR and MRAR}
The inference process of 1-Step AR is shown as the Algorithm~\ref{alg:1step}: 
\begin{algorithm}

    \caption{1-Step Autoregression inference process}
    \label{alg:1step}
    \textbf{Input:} class token [\text{C}] and mask token $\text{Mask}$\\
    \textbf{Output:}the generated image $\text{R}_\text{img}$ corresponding to the class token [\text{C}]
    \begin{algorithmic}[1]
        \STATE $\text{input} = \text{Concat}([\text{C},\text{Mask}])$\\
       \STATE $c = \text{AR Transformer(input)}$\\
       \STATE $\text{R}_\text{img} = \text{Diff}(c,\epsilon)$
    \end{algorithmic}
\end{algorithm}

The inference process of MRAR is shown as the Algorithm~\ref{alg:MRAR}:
\begin{algorithm}

    \caption{Multi-Reference Autoregression inference process}
    \label{alg:MRAR}
    \textbf{Input:} class token [\text{C}] and mask token $\text{Mask}$\\
    \textbf{Output:}the generated image $\text{R}_\text{img}$ corresponding to the class token [\text{C}]
    
    \begin{algorithmic}[1]
        \FOR{$i = 0$ to $n$}
    \IF{$i == 0$}
        \STATE $\text{input} = \text{Concat}([\text{C},\ \text{Mask}])$
    \ELSE
        \STATE $\text{input} = \text{Concat}([\text{C},\ \text{Mask},\ o_{\text{img}_0},...,o_{\text{img}_{i-1}}])$
    \ENDIF
    \STATE $c_{\text{img}_i} = \text{AR Transformer(input)}$
    \STATE $c=\text{Concat}([c_{\text{img}_0}, c_{\text{img}_1},...,c_{\text{img}_i}])$
    \STATE $o_{\text{img}_i}=\text{Diff}(c,\epsilon) $
\ENDFOR
\STATE $\text{R}_\text{img} = o_{\text{img}_n}$
    \end{algorithmic}
\end{algorithm}

\section{Implementation Details}
\label{sec: ID}
Our AR Transfomer architecture follows the Transformer~\cite{vaswani2017attention} implementation and the diffusion model architecture follows the DiT~\cite{peebles2023scalable}. TransDiff is available in three model sizes: \textbf{Huge}(AR Transformer has $40$ blocks and a width of $1280$ and diffusion decoder has $20$ blocks and a width of $1280$),\textbf{Large}(AR Transformer has $32$ blocks and a width of $1024$ and diffusion decoder has $16$ blocks and a width of $1024$), and \textbf{Base}(AR Transformer has $24$ blocks and a width of $768$ and diffusion decoder has $12$ blocks and a width of $768$).

TransDiff with 1-Step AR is trained with model sizes ranging from $290$M to $1.3$B parameters, on $32\times 8$ Nvidia A800 GPU machines with a batch size of $2048$ for $800$ epochs, using the AdamW optimizer~\cite{loshchilov2017decoupled} with a learning rate of $8.0e-4$ and a weight decay of $0.02$, along with the exponential moving average (EMA)~\cite{karras2024analyzing} strategy. And after fine-tuning with the MRAR paradigm for 40 epochs using a separate learning rate of $5.0e-5$, we obtain TransDiff with MRAR.

\section{Metric for Evaluating the Diversity of Image Semantic Features}  
In this paper, we quantify feature diversity by computing cosine similarity matrix constructed from the $256$ L2-normalized image representations. Specifically, given the feature matrix $ \mathbf{A} = [\mathbf{a}_1; \mathbf{a}_2; \dots; \mathbf{a}_{256}] \in \mathbb{R}^{256 \times d} $, where each row $ \mathbf{a}_i $ is an L2-normalized semantic feature vector, the cosine similarity between any two features $ \mathbf{a}_i $ and $ \mathbf{a}_j $ is computed as:
\begin{equation}
    \text{CosSim}(\mathbf{a}_i, \mathbf{a}_j) = \mathbf{a}_i \cdot \mathbf{a}_j
\end{equation}
The full cosine similarity matrix is then obtained via:
\begin{equation}
\mathbf{S} = \mathbf{A} \cdot \mathbf{A}^T =
\begin{bmatrix}
\mathbf{a}_1 \cdot \mathbf{a}_1 & \mathbf{a}_1 \cdot \mathbf{a}_2 & \cdots & \mathbf{a}_1 \cdot \mathbf{a}_{256} \\
\mathbf{a}_2 \cdot \mathbf{a}_1 & \mathbf{a}_2 \cdot \mathbf{a}_2 & \cdots & \mathbf{a}_2 \cdot \mathbf{a}_{256} \\
\vdots & \vdots & \ddots & \vdots \\
\mathbf{a}_{256} \cdot \mathbf{a}_1 & \mathbf{a}_{256} \cdot \mathbf{a}_2 & \cdots & \mathbf{a}_{256} \cdot \mathbf{a}_{256}
\end{bmatrix}
\end{equation}
To evaluate feature diversity while excluding self-similarities, we define:
\begin{equation}
\text{Cosine Similarity} = \left\| \mathbf{S}\right\|_{L1}
\end{equation}
A smaller cosine similarity score indicates that the generated image representations are more diverse in the semantic space.

\section{Properties of MRAR} 
\begin{figure}
    \centering
    \includegraphics[width=0.90\linewidth]{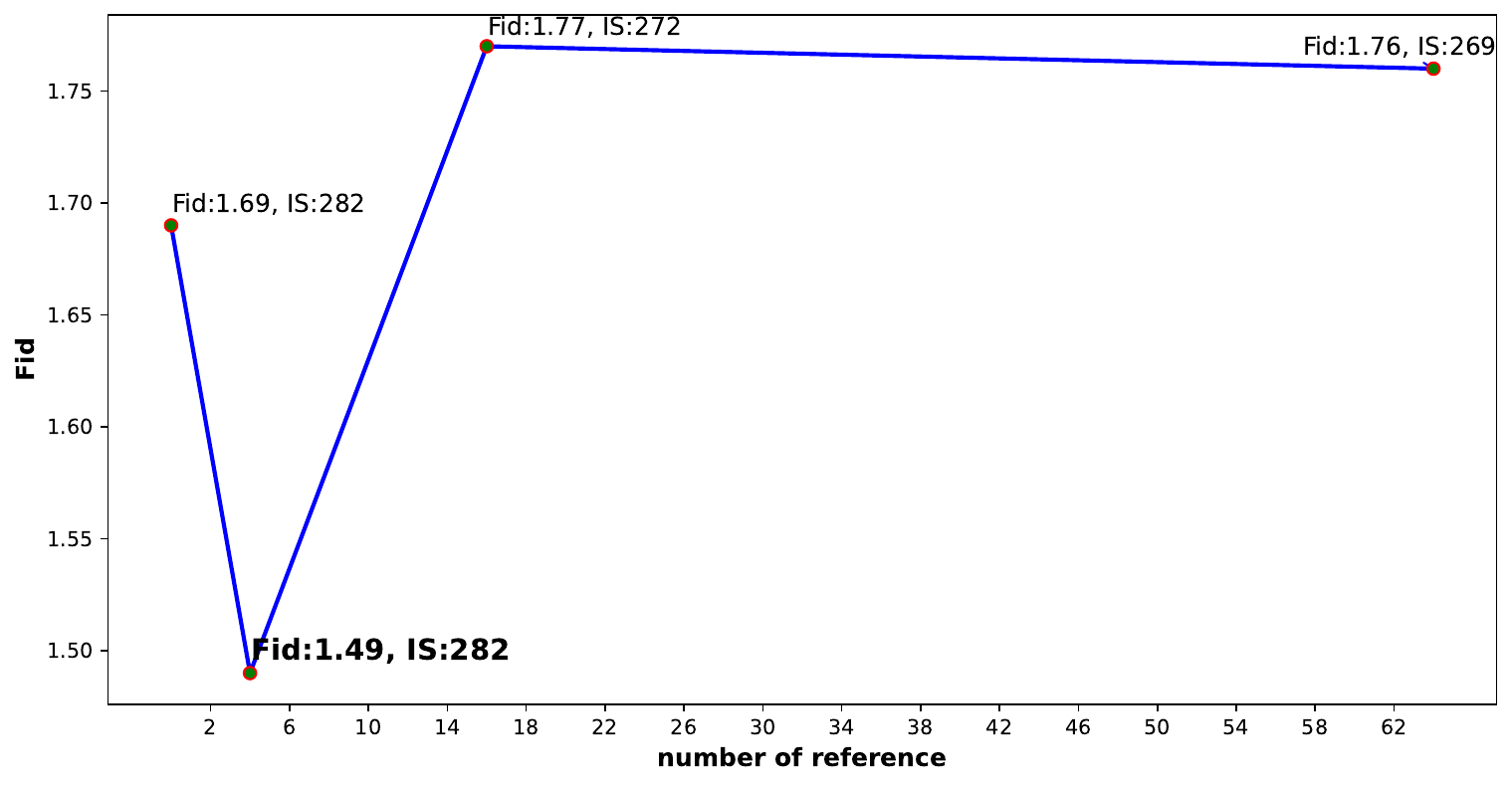}
    \caption{Correlation between the number of references and FID score.}
    \label{fig:Properties of MRAR}
\end{figure}

We extend TransDiff-L into a multi-reference input framework to explore the extended properties of MRAR. 
We systematically evaluate model performance under different reference configurations (64, 16, 4, and 0), with a particular focus on analyzing the correlation between the number of references and overall image quality (\emph{i.e.}, fid value). The Figure~\ref{fig:Properties of MRAR} shows the optimal number of references (\emph{i.e.}, $4$).

\section{The Euler–Maruyama Method}
\label{App:Euler–Maruyama}
Our diffusion module inherits the Euler–Maruyama method for image generation from noise through stochastic sampling. However, in contrast to Euler–Maruyama, we  introduce two scaling factors, $s_1$ and $s_2$, which respectively modulate the drift term and diffusion term. These parameters primarily function to optimize the generation process by balancing numerical stability with output quality through strategic term scaling. The following analytical breakdown elaborates on their operational mechanisms.

The reverse process of diffusion models is typically implemented by solving a Stochastic Differential Equation (SDE), governed by:
\begin{equation}
\label{SDE}
 dx = \left[ \mathbf{v}_\theta(x, t) - \frac{1}{2} \sigma(t)^2 \nabla_x \log p(x|t) \right] dt + \sigma(t) dW
\end{equation}
where $\mathbf{v}_\theta(x, t)$ denotes the noise predicted by the velocity model, $\sigma(t)$ represents the diffusion coefficient, governing the intensity of noise injection, $dW$ indicates the increment of a Wiener process (Brownian motion), characterizing stochastic noise.

We formally define the drift term as $d_{\text{cur}} = \mathbf{v}_\theta(x, t) - \frac{1}{2} \sigma(t)^2 s_\theta(x, t)$ and the diffusion term as $\sqrt{\sigma(t)} \cdot \Delta W$~\cite{kloeden2013numerical}. Through the Euler–Maruyama Method, the reverse process of the diffusion model can be discretized as:
\begin{equation}
\label{Euler–Maruyama}
 x_{\text{next}} = x_{\text{cur}} +  d_{\text{cur}} \cdot dt +  \sqrt{\sigma(t)} \cdot \Delta W
\end{equation}

In Eq~\ref{Euler–Maruyama}, the weights of the drift and diffusion terms are intrinsically determined by the model architecture. However, practical implementations of the vanilla Euler–Maruyama Method~\cite{bally1995euler} may encounter two critical challenges: Numerical instability and Non-trivial trade-offs.

To address these challenges, we introduce two scaling factors ($s_1$ and $s_2$) into the vanilla Euler–Maruyama framework, reformulating the reverse diffusion process as:
\begin{equation}
\label{Scale_Euler–Maruyama}
 x_{\text{next}} = x_{\text{cur}} +  s_1 \cdot d_{\text{cur}} \cdot dt +  s_2 \cdot \sqrt{\sigma(t)} \cdot \Delta W
\end{equation}

The $s_1$ factor modulates the drift term intensity by scaling the directional update step size \(d_{\text{cur}} \cdot \Delta t\). A value below 1.0 suppresses excessive drift updates to enhance numerical stability, whereas values exceeding 1.0 accelerate generative convergence at potential costs to fine-grained detail preservation.  

The $s_2$ factor regulates the diffusion term strength through scaled noise injection \(\sqrt{\sigma(t)} \cdot \Delta W\). Reducing this factor below 1.0 diminishes stochastic perturbations to improve output sharpness, while increasing it beyond 1.0 promotes sample diversity through amplified noise – a trade-off that may introduce artifacts at higher scaling magnitudes. 

\section{More Quantitative Results}

We also report our FID, IS on ImageNet $512\times 512$ benchmark,  as detailed in Tab~\ref{fig: 512_benchmark}. 

\begin{table}
\centering

\begin{tabular}{c|c|c|cc|c}
\hline
\textbf{Type} & \textbf{Model}    & \textbf{\#params} & \textbf{FID ↓} & \textbf{IS ↑} & \textbf{Time ↓} \\ \hline
Diffusion     & ADM & 554M &23.24          & 101.0         & -               \\
Diffusion     & DiT-XL/2       &  657M &3.04           & 240.8         & 81s              \\ \hline
AR            & VQGAN     & 227M &26.52          & 290.5         & 25s              \\
VAR           & VAR-d36-s      &  2.3B &2.63           & \textbf{303.2}& 1s               \\ \hline
TransDiff     & TransDiff-L, MRAR &  683M &\textbf{2.51}& 276.6& 1s               \\ \hline
\end{tabular}

\caption{ImageNet $512\times 512$ conditional generation.}
\label{fig: 512_benchmark}
\end{table}

\section{More Qualitative Results}

\begin{figure*}
    \centering
    \includegraphics[width=1.0\linewidth]{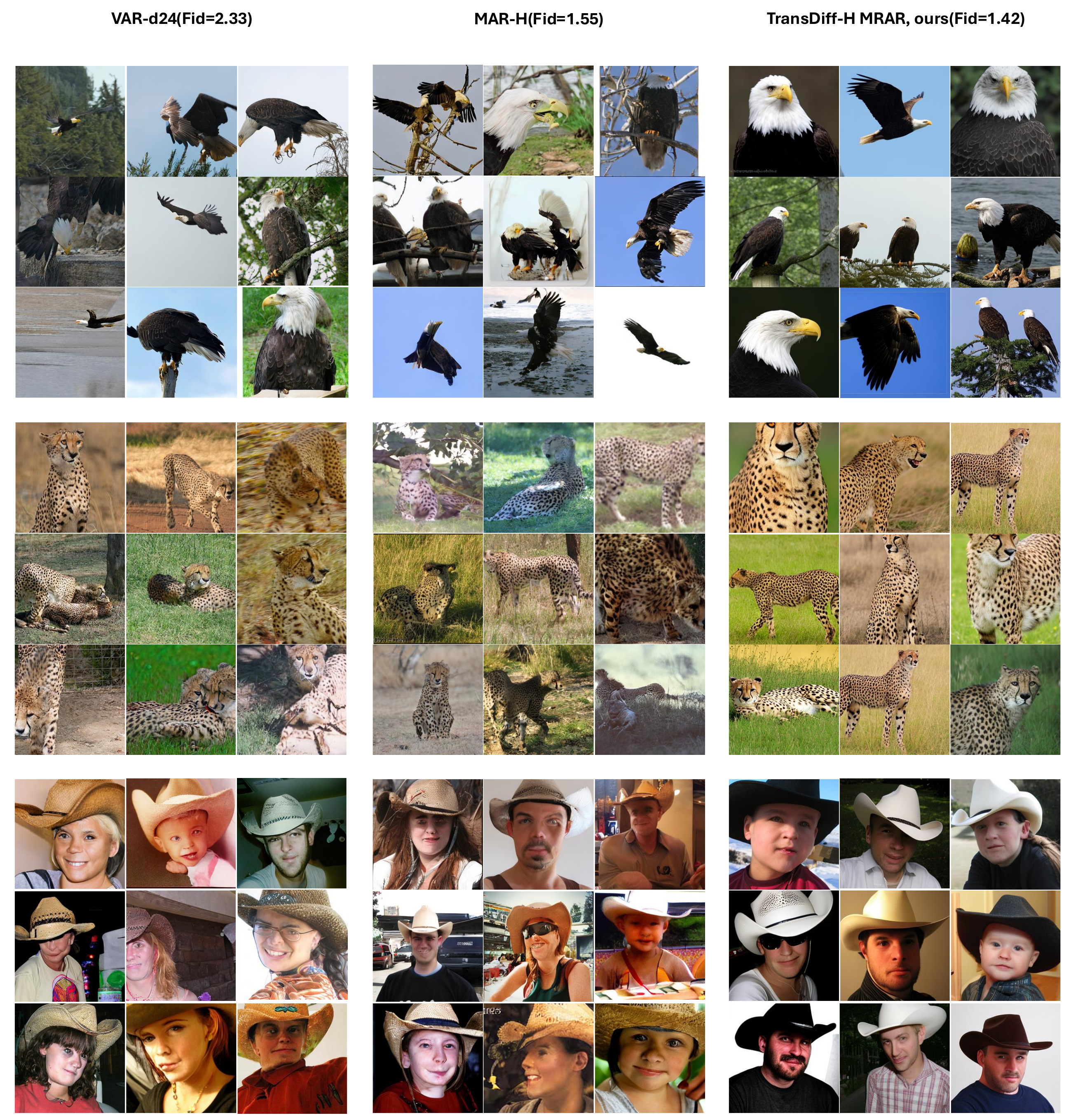}
    \caption{Model comparison on ImageNet $256\times 256$ benchmark.}
    \label{fig: 256}
\end{figure*}

\begin{figure*}[p]
    \centering
    \includegraphics[width=0.9\linewidth]{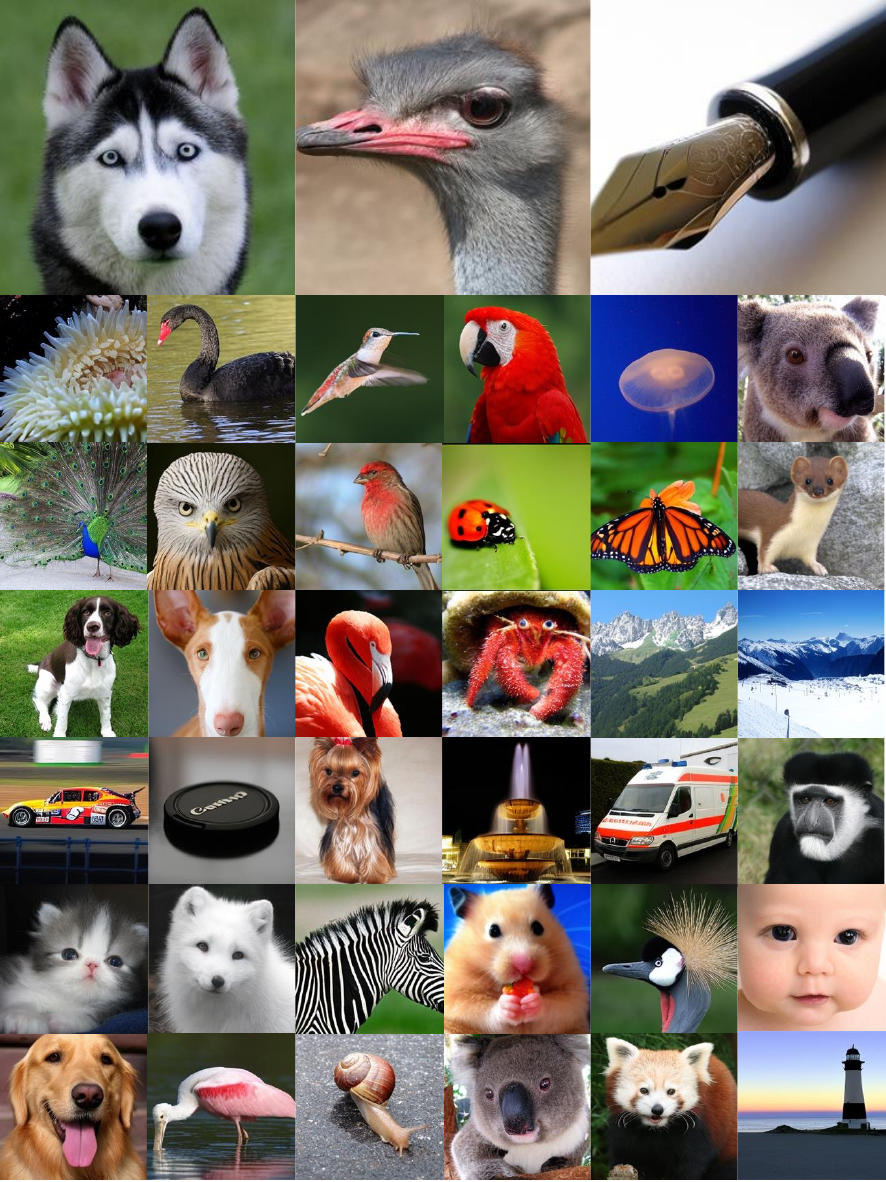}
    \caption{\textbf{Qualitative Results}. More $256\times 256$ class-conditional generation results from TransDiff-H, MRAR on ImageNet.}
    \label{fig: 256_demo}
\end{figure*}

\begin{figure*}[p]
    \centering
    \includegraphics[width=0.9\linewidth]{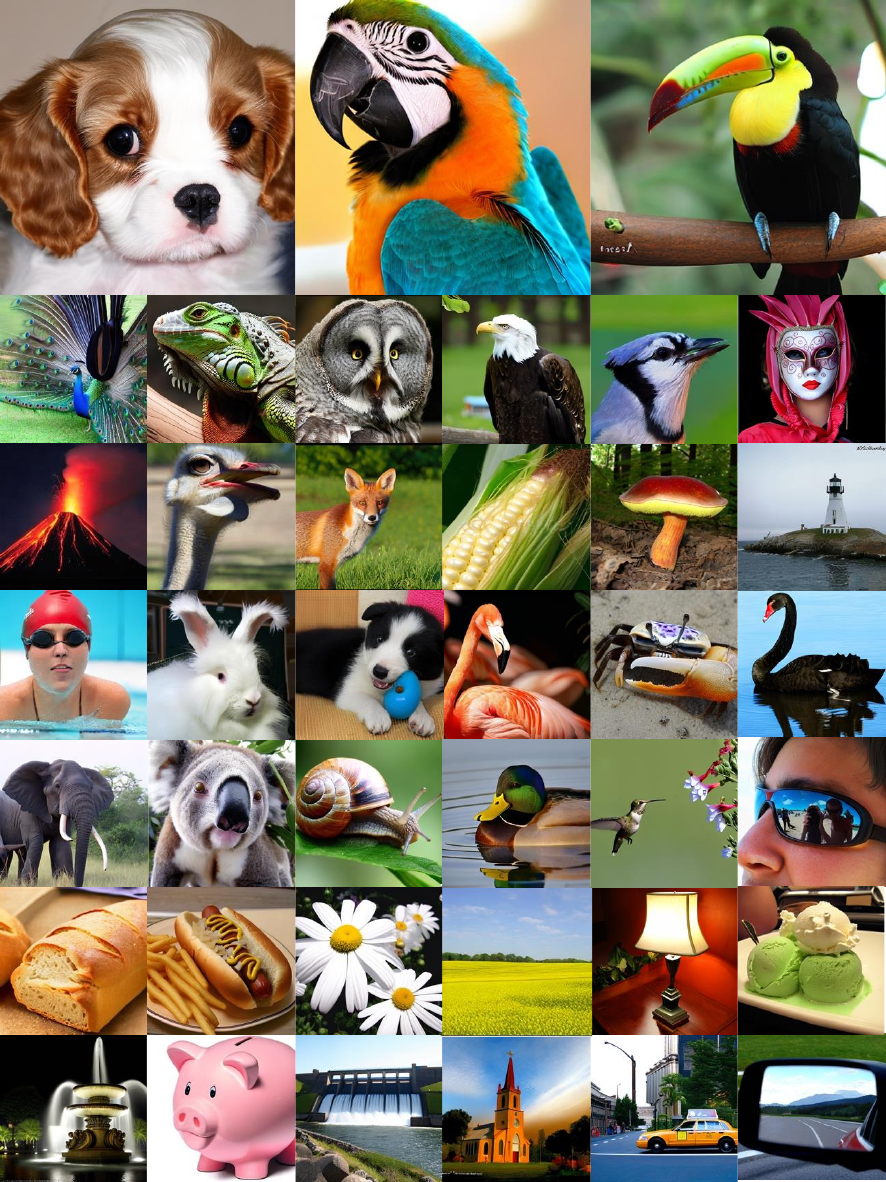}
    \caption{\textbf{Qualitative Results}. More $512\times 512$ class-conditional generation results from TransDiff-L, MRAR on ImageNet.}
    \label{fig: 512_demo}
\end{figure*}

We also provide more visualization examples in Figure~\ref{fig: 256},~\ref{fig: 256_demo} and~\ref{fig: 512_demo}. As presented, Transdiff-H MRAR produces high-quality, diverse, and content-rich image samples. Furthermore, as illustrated in Figure~\ref{fig: 256}, compared to VAR and MAR under similar parameter settings, 
Transdiff produces more realistic textures, sharper facial features, and consistent lighting. This highlights its strong potential for complex semantic tasks like human face synthesis.

\end{document}